\pdfoutput=1

\documentclass[11pt]{article}

\usepackage[final]{acl}

\usepackage{times}
\usepackage{latexsym}
\usepackage{setspace}
\usepackage{multirow}
\usepackage{floatrow}
\usepackage{amsmath}
\usepackage{amssymb}
\usepackage{booktabs}
\usepackage{enumitem}
\usepackage{pifont}
\usepackage{url}

\usepackage[T1]{fontenc}

\usepackage[utf8]{inputenc}

\usepackage{microtype}

\usepackage{inconsolata}

\usepackage{graphicx}

\title{CADReview: Automatically Reviewing CAD Programs with Error Detection and Correction}

\author{
 \textbf{Jiali Chen\textsuperscript{1,2}\thanks{Equal Contribution.}},
 \textbf{Xusen Hei\textsuperscript{1,2}$^*$},
 \textbf{Hongfei Liu\textsuperscript{1,2}},
  \textbf{Yuancheng Wei\textsuperscript{1,2}},
 \textbf{Zikun Deng\textsuperscript{1,2}}, \\
 \textbf{Jiayuan Xie\textsuperscript{3}},
 \textbf{Yi Cai\textsuperscript{2,1}$^\dagger$},
  \textbf{Qing Li\textsuperscript{3}}
\\
 \textsuperscript{1}School of Software Engineering, South China University of Technology, \\
\textsuperscript{2}Key Laboratory of Big Data and Intelligent Robot Ministry of Education, \\
 \textsuperscript{3}The Hong Kong Polytechnic
University
\\
 \texttt{segarychen@mail.scut.edu.cn} \quad \texttt{\{zkdeng, ycai\}@scut.edu.cn} \\
\texttt{\{jiayuan.xie, qing-prof.li\}@polyu.edu.hk}\\
 \small{
   \textbf{$^\dagger$Correspondence:} \href{mailto:ycai@scut.edu.cn}{ycai@scut.edu.cn}
 }
}

\newcommand{\ie}{\emph{i.e.}}
\newcommand{\eg}{\emph{e.g.}}

\newcommand{\etc}{\emph{etc}}

\begin{document}
\maketitle
\begin{abstract}
Computer-aided design (CAD) is crucial in prototyping 3D objects through geometric instructions (\ie, CAD programs).
In practical design workflows, designers often engage in time-consuming reviews and refinements of these prototypes by comparing them with reference images.
To bridge this gap, we introduce the CAD review task to automatically detect and correct potential errors, ensuring consistency between the constructed 3D objects and reference images.
However, recent advanced multimodal large language models (MLLMs) struggle to recognize multiple geometric components and perform spatial geometric operations within the CAD program, leading to inaccurate reviews.
In this paper, we propose the CAD program repairer (ReCAD)
framework to effectively detect program errors and provide helpful feedback on error correction. Additionally, we create a dataset, \textit{CADReview}, consisting of over 20\textit{K} program-image pairs, with diverse errors for the CAD review task. 
Extensive experiments demonstrate that our ReCAD significantly outperforms existing MLLMs, which shows great potential in design applications\footnote{Our dataset and code are released at the project page: \url{https://cgl-pro.github.io/cadreview}}.

\end{abstract}

\section{Introduction}
Computer-aided design (CAD) plays a critical role in industrial design and manufacturing, serving as the foundation for product prototyping ~\cite{deepcad,shapecoder}.
It represents 3D objects through sequences of geometric instructions, commonly referred to as \textit{CAD programs}, which 
define editable geometric components and operations.
Despite the emergence of various 3D modeling software (\eg, AutoCAD, SketchUp, Rhino and FreeCAD), the design workflow persists as a technically challenging process. It is time-consuming and requires specialized expertise from designers.

\begin{figure}[!]
  \centering
  \includegraphics[scale=0.505]{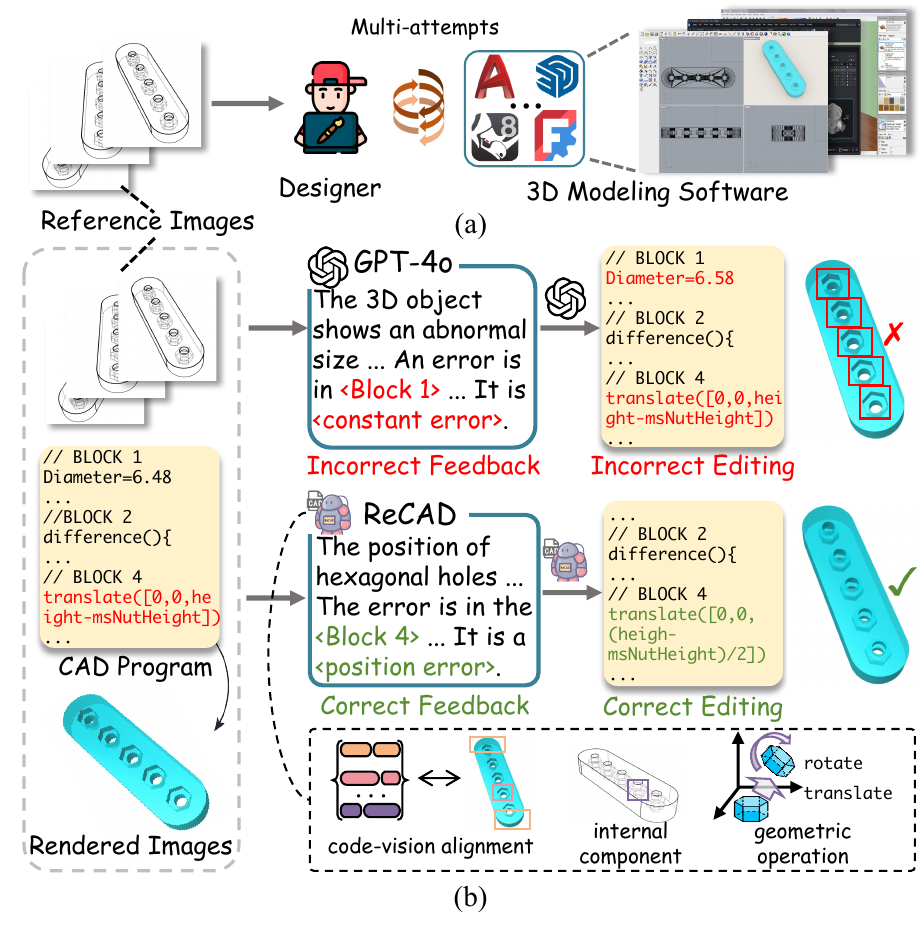}
  \caption{Overview of different ways (\ie, designer, GPT-4o and our ReCAD) to CAD review. The \textcolor[RGB]{88,142,49}{green} and \textcolor[RGB]{255,0,0}{red} denote correct and incorrect prediction, respectively.}
\label{pic_intro_case}
\end{figure}

Towards this end, several generative models have emerged for 3D object generation through CAD programs ~\cite{deepcad,skexgen,cad-signet}.
However, the generated 3D object may contain errors, and the practical requirements in industrial design extend far beyond mere initial 3D generation.
In realistic design workflows, designers conduct 
meticulous reviews and refine the 3D object using modeling software \cite{cad-book-1}, a process that requires multiple attempts to ensure consistency with the design drawing (\ie, reference image), as shown in Fig. \ref{pic_intro_case} (a).
In contrast, AI models can directly analyze and edit CAD programs for this process. 
Motivated by this intuition, we propose the CAD review task to automatically detect and correct program errors, addressing discrepancies between the 3D object and reference images.
Following previous works \cite{coffee,codeagent} on code review, our task aims to generate helpful feedback for program error detection and then utilize this feedback to edit the erroneous code,
as illustrated in Fig. \ref{pic_intro_case} (b).

Currently, advanced multimodal large language models (MLLMs), \eg, GPT-4o \cite{gpt4o}, have shown remarkable performance across vision-language tasks ~\cite{mmmu,l2c,fire,yuan2023joint,yuan2025}. 
Nevertheless, they struggle to integrate CAD programs with reference images for the CAD review task. 
As illustrated in Fig. \ref{pic_intro_case} (b), GPT-4o fails to detect program errors, primarily due to its limited ability to align code with visual information.
Therefore, addressing the CAD review task requires overcoming multiple challenges.
\textbf{First}, 3D objects typically consist of multiple components, each represented by a specific code block within the program.
Moreover, these code blocks may contain intricate program structures (\eg, subroutines and control flows), which recursively organize primitives (\eg, cuboid and ellipsoid) into geometric components. 
The model demands correlating these components with corresponding code blocks.
\textbf{Second}, 3D objects may contain hidden internal components that hinder visual inspection for humans and models, \eg, internal hexagonal holes shown in Fig. \ref{pic_intro_case} (b).
It requires programmatic analysis of the CAD programs for error detection.
\textbf{Third}, in addition to error detection, precise correction requires mapping geometric operations to corresponding code modifications (\eg, rotation and translation). As shown in Fig. \ref{pic_intro_case} (b), it needs to modify the component's offset along the z-axis.

To tackle the above challenges, 
we propose CAD Program Repairer (ReCAD), a MLLM-based framework, consisting of a feedback generator and code editor.
Specifically, the feedback generator first aligns code with geometric components and then detects program errors for accurate feedback generation.
Next, the code editor leverages this feedback to perform geometric operations by editing code for 3D object reconstruction.
Moreover,
we create \textit{CADReview}, a benchmark dataset for the CAD review task, consisting of over 20\textit{K} reference images, CAD programs, and annotated feedback. It contains 3D objects, each averaging over 8 geometric components, meeting the realistic industrial design requirements.

Our contributions are summarized as follows:
\textbf{(i)} Motivated by the realistic design workflows, we introduce the CAD review task, which aims to detect and correct CAD program errors based on the reference images.
\textbf{(ii)} We develop ReCAD, a MLLM-based framework that generates feedback to guide program error correction. Additionally, we create \textit{CADReview} dataset as a new benchmark for the CAD review task. 
\textbf{(iii)} Experimental results demonstrate that our ReCAD achieves significant performance gains over existing MLLMs.

\section{Related Work}
\noindent
\textbf{CAD Program.}
Designing CAD programs is inherently challenging, as each code block on the final geometric output is often non-intuitive and difficult to foresee. 
Existing research \cite{deepcad,skexgen,shapewalk} primarily focuses on using AI models to generate CAD programs. 
Specifically, DeepCAD \cite{deepcad} and SkexGen \cite{skexgen} are 3D generative models that represent shapes with CAD programs, offering a novel perspective on 3D representation. 
Text2CAD \cite{text2cad} generates CAD programs based on textual modeling instructions, with diverse levels ranging from beginner to expert.
Recent studies \cite{cadvlm,cad-gpt} have explored using multimodal large language models (MLLMs) for CAD program generation.
Specifically, CAD-GPT \cite{cad-gpt} generates CAD programs based on multimodal input,
which maps 3D spatial coordinates to a 1D linguistic feature space through tokenization method.
Despite advancements, the generated CAD programs still exhibit discrepancies from the intended design.
Since CAD designers spend considerable time reviewing 3D objects against design drawings, we introduce the CAD review task for automatic program error detection and correction. 
Furthermore, we also propose the \textit{CADReview} dataset, which includes diverse geometric components and complex structures for review.

\noindent
\textbf{Code Edit.}
Large language models (LLMs) have achieved impressive results in code editing on generic programming languages (\eg, C++ and Python).
A common approach for code editing involves utilizing execution feedback from compilers or test cases \cite{code-self-edit,self-debug,critic} to correct code.
Recently, natural language feedback has been utilized for code editing due to its interpretability.
Specifically, \citet{coffee} design a reward function that reflects the helpfulness of feedback.
CodeAgent ~\cite{codeagent} is an autonomous multi-agent system, which simulates collaboration among roles in practical software development to perform code editing.
However, CAD programs fundamentally differ from generic programs, as their code blocks represent geometric components, and the editing process relies on design drawings.
To the best of our knowledge, we are the first to introduce the task of CAD code editing, termed CAD review.

\section{Building \textit{CADReview} Dataset}
While a few datasets of CAD programs have been recently proposed \cite{deepcad,cadtalk}, they do not include code segments that deviate from the design drawing (\ie, reference image), which are critical for designers to rectify.
Hence, we introduce \textit{CADReview} dataset as a new benchmark for the proposed CAD review task. 
In total, we split 17,334, 2,000 and 1,615 data samples for training, validation and testing respectively.
Our dataset contains CAD programs with potential errors and corresponding correct reference images.
In the following sections, we describe the construction process, as shown in Fig. \ref{pic_dataset_cons}. 
More details are provided in Appendix \ref{app_dataset}.

\subsection{Collecting Program-Image Pairs} \label{dataset-program}
We collect two types of CAD programs (\ie, human-made and machine-made programs), and 
obtain the corresponding design drawings as reference images.
Human-made programs involve complex geometric instructions, \eg, control flow, while machine-made programs are constructed from simple geometric primitives.
It enables a comprehensive evaluation of program structures and geometric complexity.
We choose OpenSCAD ~\cite{openscad}, an open-source CAD program language for its convenience in data collection.

\begin{figure}[!]
  \centering
  \includegraphics[scale=0.331]{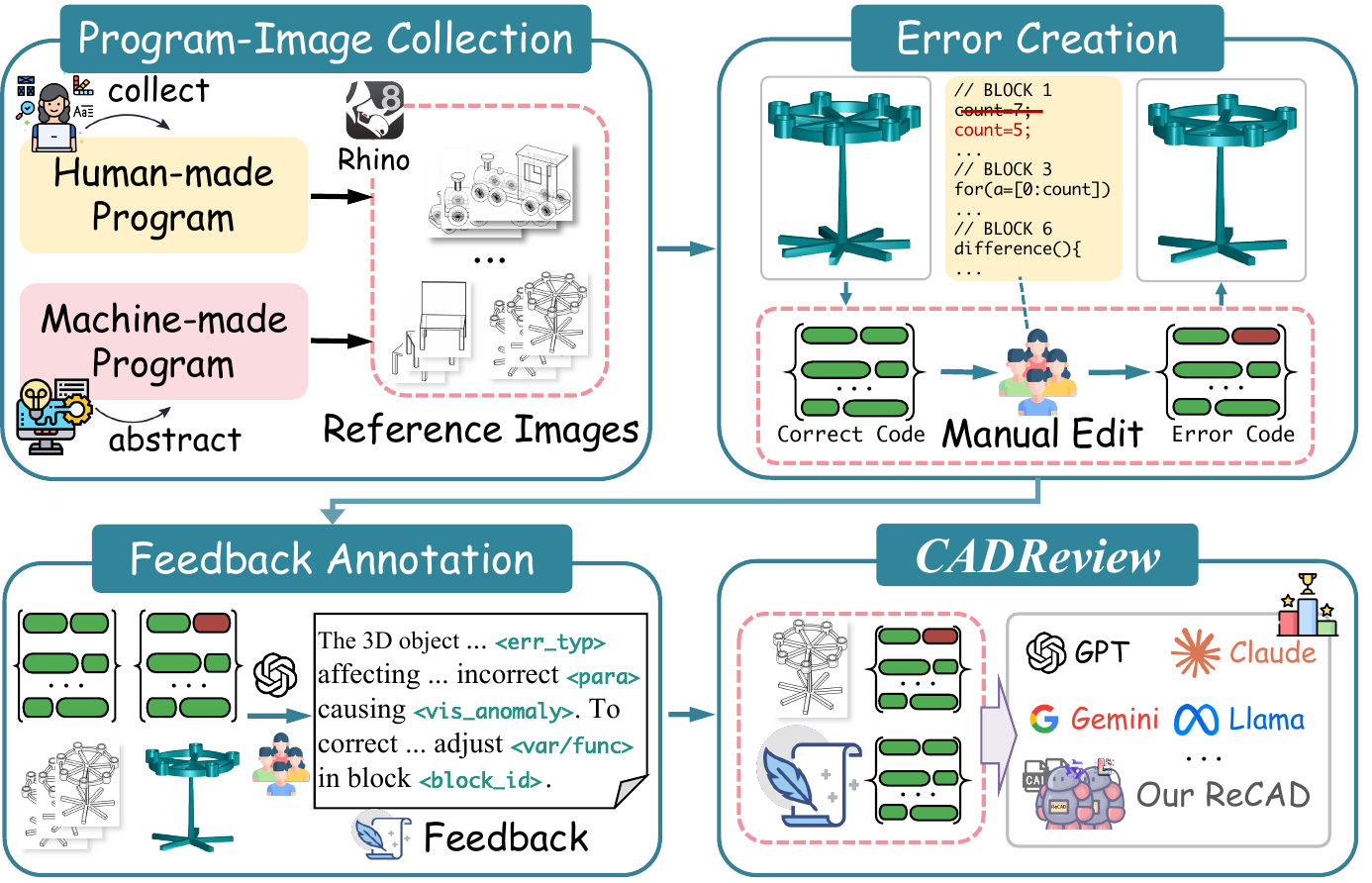}
  \caption{Construction workflow of our \textit{CADReview} dataset, including program-drawing pairs collection, error creation and feedback annotation.}
\label{pic_dataset_cons}
\end{figure}

\vspace{0.5em}
\noindent
\textbf{Human-made Program.}
%
The human-made programs are created by experienced designers. We collect and filter 1.5\textit{K} OpenSCAD programs from online design communities,
with diverse 3D object categories, as shown in Fig. \ref{pic_data_demo} of Appendix \ref{app_data_overview}.
Specifically, we first render the corresponding 3D objects for each collected program.
Considering that designers often divide 3D objects into distinct geometric components for review, we parse the CAD program into multiple code blocks as different components.
To achieve this, we traverse the abstract syntax tree of the program from the top down. During traversal, we treat the irreducible parts, \ie, macros, modules, control flows (\eg, nested loops and conditional statements), boolean operations (\eg, difference, union, intersection, \etc), and geometric primitives (\eg, cuboid, ellipsoid, \etc) as independent code blocks. 
For simplicity, we comment block IDs before the corresponding code blocks and treat initial macros as the first block.

\vspace{0.5em}
\noindent\textbf{Machine-made Program.}
Previous studies \cite{shapecoder,cadtalk} have shown that 3D objects can be automatically abstracted as compositions of simple geometric primitives.
Building on this insight, we choose three basic object categories (\ie, chair, table and storage) from PartNet ~\cite{partnet} and convert them into OpenSCAD programs with cuboid, ellipsoid and cylinder abstraction.
Specifically, we use cuboids for the initial abstraction and then randomly replace some cuboids with ellipsoids or cylinders. 
After abstraction, we manually remove low-quality samples with excessive overlap of primitives.
Similar to human-made programs, we also separate code blocks and assign corresponding block IDs. 
As a result, we obtain a total of 1.6\textit{K} machine-made programs.

\vspace{0.5em}
\noindent\textbf{Reference Image.}
For each CAD program, we perform multiview rendering to obtain three representative design drawings of the 3D object as reference images.
Specifically, we use the modeling software (\ie, Rhino ~\cite{rhino}) for rendering, as shown in Fig. \ref{pic_dataset_cons}. 
The viewpoints are manually sampled from a hemisphere around the object. 
In particular, we utilize a perspective view to render objects considering their potential internal components.

\begin{figure*}[!]
  \centering
  \includegraphics[scale=0.63]{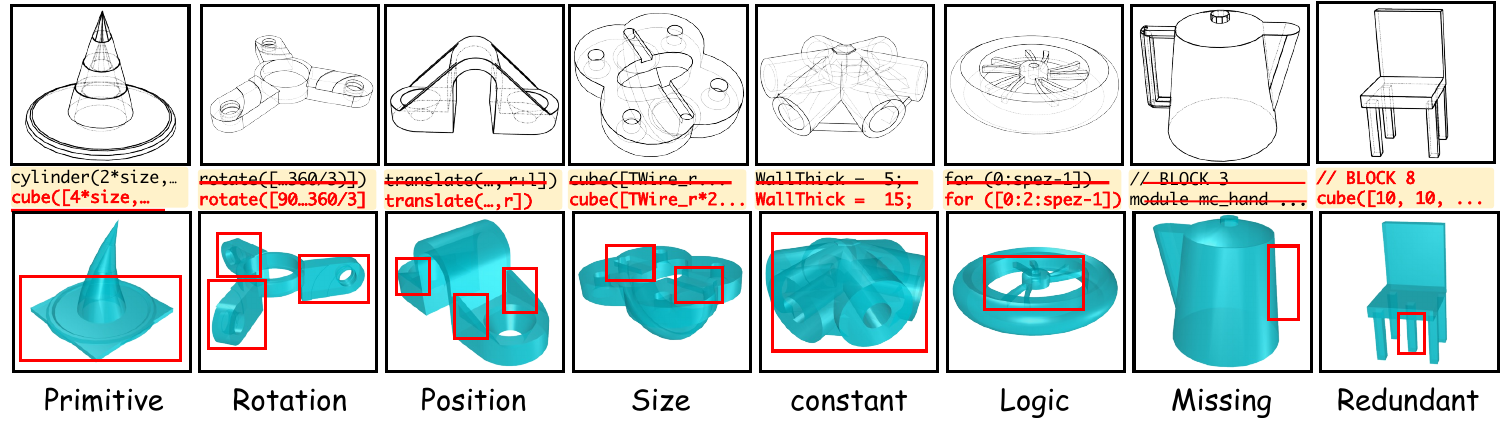}
  \caption{Examples of 8 types of CAD program error from our \textit{CADReview} dataset. \textbf{Top row}: The reference images from one sampled viewpoint.
  \textbf{Second row}: Error creation on CAD programs, showing the edited code segments. 
  \textbf{Bottom row}: The rendered 3D objects by erroneous programs.
  }
  \label{pic_error_types}
\end{figure*}

\subsection{Creating Errors on CAD Programs}
We manually modify programs to create errors, resulting in anomalous 3D objects that exhibit discrepancies with the reference images.
Our dataset includes 8 error scenarios that are relevant to real-world CAD review applications, as shown in Fig. \ref{pic_error_types}.
\textit{(i) Primitive error}
refers to the use of an incorrect geometric primitive. For instance, substituting a cube with a cylinder or a sphere leads to a misrepresentation of the intended design.
\textit{(ii) Rotation error} are created by applying a rotational transformation to the component of the 3D object.
\textit{(iii) Position error} pertains to deviations in the 3D coordinates of the components from their intended positions in the design.
\textit{(iv) Size error} is the discrepancy in the scale of the geometric component, similar to scenarios where parts of the object appear broken.
\textit{(v) Constant error} represents the errors in the initial macros or constants. These incorrect values or invalid assignments disrupt the intended 3D object generation.
\textit{(vi) Logic error} occurs in control flow statements, such as logical conditionals and loops, that cause unintended program execution paths. 
\textit{(vii) Missing block} remove one code block, which results in incomplete 3D object construction or missing components.
\textit{(viii) Redundant block} involves the unnecessary code blocks that do not contribute to the intended 3D object.

We apply these errors to both human-made and machine-made CAD programs. Specifically, we develop a web interface to manually create errors on the CAD programs.
Each annotator is assigned to create one error per annotation. A sourced program can be converted into multiple erroneous programs, where errors may appear in different code blocks.
Additionally, normal samples with correct programs are also included in our dataset.

\subsection{Annotating Feedback on CAD Programs}
Inspired by previous studies \cite{coffee} leveraging natural language feedback for code editing, we use GPT-4o \cite{gpt4o} to annotate feedback for CAD programs.
Our feedback focuses on evaluating the consistency of the CAD program with the reference images.
The feedback provides a textual description of the visual anomalies, erroneous code blocks, and types of program errors.
Specifically, for erroneous CAD programs, we first prompt GPT-4o to compare the reference images with the rendered output image, highlighting the differences and providing descriptions of the visual anomalies. 
Then, GPT-4o needs to identify the block IDs of the incorrect code and describe the errors based on the original correct code.
After that, we manually recheck each feedback to ensure accuracy.
For correct CAD programs, we predefine feedback in Table \ref{tab_cor_feedback}.

\begin{figure*}[]
  \centering
  \includegraphics[scale=0.535]{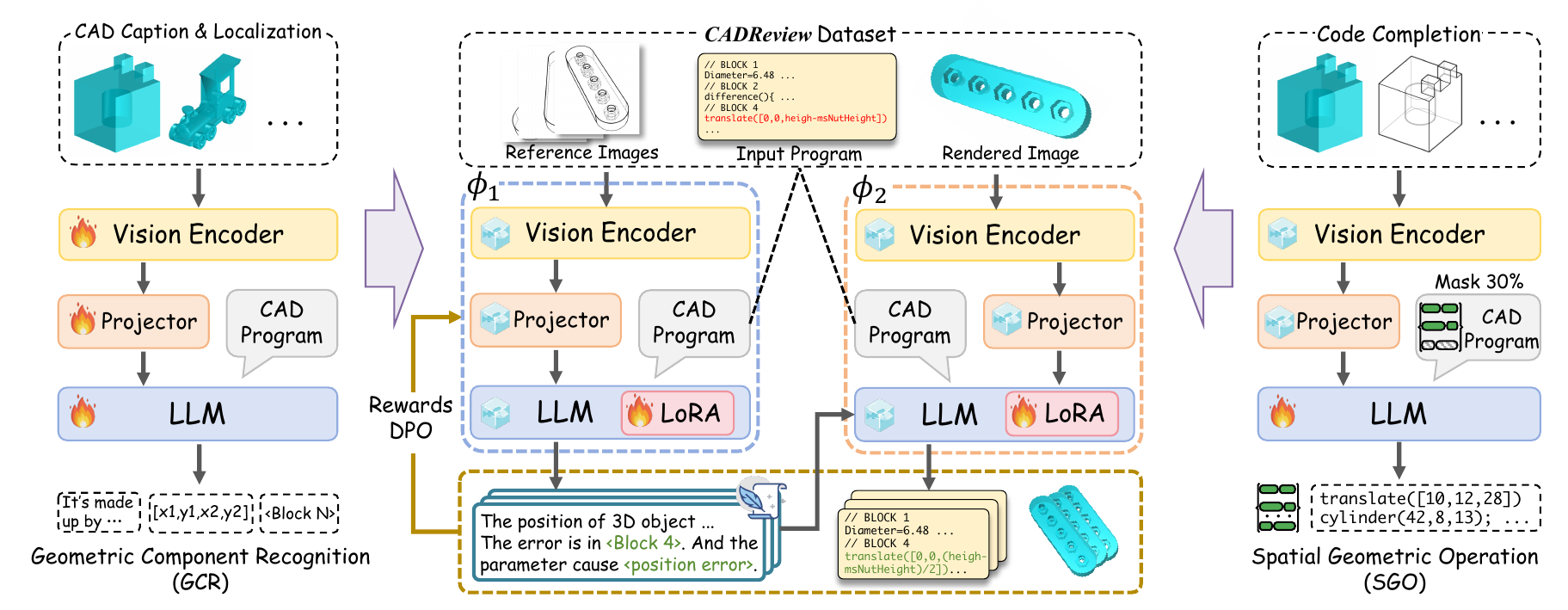}
  \caption{Overview of our 
  ReCAD. We design geometric component recognition (GCR) and spatial geometric operation (SGO) mechanisms to initialize feedback generator $\phi_1$ and code editor $\phi_2$, respectively.
  } 
  \label{pic_framework}
\end{figure*}

\section{Proposed Method}
Given the reference image and the potentially inconsistent CAD program, our goal is to detect and correct program errors. 
It requires recognizing geometric components and performing spatial geometric operations within the CAD program.
To this end, we propose a multimodal large language model (MLLM)-based framework, CAD Program Repairer (ReCAD), which generates helpful feedback for program error detection and then utilizes the feedback for error correction, as illustrated in Fig. \ref{pic_framework}.
Specifically, we build ReCAD by combining the feedback generator $\phi_1$ and code editor $\phi_2$ with the same MLLM backbone, where $\phi_1$ produces descriptions of errors in the CAD program as feedback, offering guidance for $\phi_2$ to correct.
We train them with two-stage supervised fine-tuning (SFT), followed by reinforcement learning (RL) with reward functions.
Each MLLM consists of three key modules: vision encoder, large language model (LLM) and vision-language projector.
Moreover, we also explore different MLLMs (\ie, Qwen2-VL \cite{qwen2-vl} and LLaVA-OV \cite{llava-ov}) as the backbone for both $\phi_1$ and $\phi_2$. 
In the following, we 
elaborate on the details of our ReCAD framework.

\subsection{Feedback Generator}
Feedback on the CAD program aims to 
describe potential discrepancies between the program and reference images.
Although existing models can generate feedback for traditional code editing \cite{coffee}, they struggle to align geometric components with corresponding code blocks in the CAD domain, leading to ineffective feedback. 
Therefore, we propose the geometric component recognition (GCR) mechanism to enhance the ability of feedback generator $\phi_1$ to recognize both visual and programmatic components.

\subsubsection{Geometric Component Recognition}
We collect and augment 
over 700\textit{K} CAD caption and localization data to train $\phi_1$ for GCR.
More details about the data are provided in Appendix \ref{app_pretrain}.

\vspace{0.5em}
\noindent
\textbf{CAD Caption.} 
In contrast to traditional image caption, CAD caption aims to textually describe the components that make up the 3D object, rather than the entire object.
Specifically, we use images and beginner-level textual CAD modeling instructions from the Text2CAD \cite{text2cad} dataset as image-text pairs for captioning. 
The instruction primarily includes shape properties, 
\eg, a circular CAD model with a central hole, providing guidance for MLLM to visually recognize geometric components.
Following \cite{llava-1-5}, we only train the vision-language projector while keeping the vision encoder and large language model (LLM) frozen.
After that, 
the projector effectively refines visual features, enabling the LLM to focus on these geometric components.

\begin{figure}[!]
  \centering
  \includegraphics[scale=0.455]{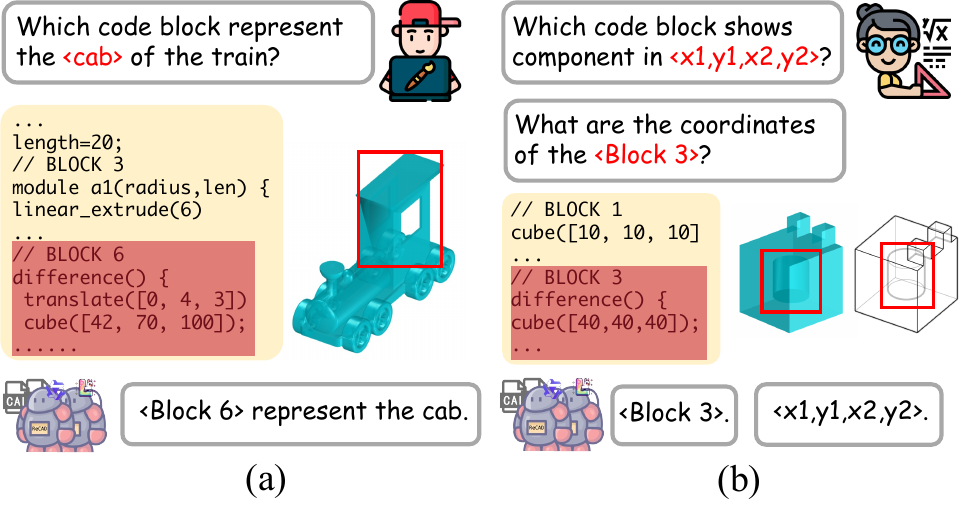}
  \caption{Overview of CAD localization. (a) Semantic matching. (b) Coordinate matching and localization.}
\label{pic_cad_loc}
\end{figure}

\vspace{0.5em}
\noindent
\textbf{CAD Localization.} 
CAD localization aims to align the recognized geometric components with the CAD program.
It can be divided into three aspects, as illustrated in Fig. \ref{pic_cad_loc}.
\textit{(i) Semantic matching} predict the code block based on the corresponding specific semantic labels.
\textit{(ii) Coordinate matching} queries which code block corresponds to the specific coordinates of the rendered image.
\textit{(iii) Coordinate localization} determines the coordinates of a given code block.
To build the training data, we first utilize the CADTalk \cite{cadtalk} dataset, which provides a semantic label for each code block, 
\eg, cab of the train in Fig. \ref{pic_cad_loc} (a).
Since the CAD review task involves evaluating internal components, we define rules and prompt GPT-4o to generate CAD programs comprising basic primitives and boolean operations (\eg, difference and union), as shown in Fig. \ref{pic_cad_loc} (b).
In particular, we collect two types of images to assist the model in simultaneously understanding both the reference and rendered images for CAD review.
Inspired by the continual pre-training \cite{intern-vl-2-5}, we initialize the $\phi_1$ trained on the CAD caption and 
jointly fine-tune the vision encoder and LLM for CAD localization.
After sequential training, 
we observe that localization performance achieves high accuracy at nearly 90\%, 
while the performance of CAD caption remains unaffected.
\subsubsection{SFT for Feedback Generation} 
After sequential training on CAD caption and localization, we perform supervised fine-tuning (SFT) of $\phi_1$ 
to generate feedback. 
Specifically, for each input CAD program, 
we first obtain the rendered image from a specific viewpoint as supplementary input. 
The reference images, CAD program and rendered image are then fed into frozen $\phi_1$ with learnable LoRA \cite{lora} layers for language modeling. 
Lastly, we use the cross-entropy loss to maximize the probability of predicting ground truth feedback.

\subsection{Code Editor}
In addition to recognizing geometric components, the code editor $\phi_2$ utilizes the generated feedback as guidance to perform geometric operations for program error correction.
We propose the spatial geometric operation (SGO) mechanism for the code editor $\phi_2$ to learn spatial relationships and geometric transformations within the CAD programs. 

\subsubsection{Spatial Geometric Operation}
The CAD program mainly consists of operators (\eg, rotation, translation, boolean operations) and operands (\eg, displacement along axes, angles). 
Due to the lack of explicit scalar values in the input image, the code editor $\phi_2$ learns these operations through code completion, where input code provides the necessary reference spatial context. 
Specifically, we use the same data sources as CAD localization (\ie, CADTalk and LLM-augmented data) and randomly mask 30\% code blocks for $\phi_2$ to predict.
To maintain the geometric component recognition ability, we initialize $\phi_2$ by the MLLM trained on the CAD caption and localization.

Through empirical observation, we find that the cross-entropy loss converges rapidly to a minimal value when $\phi_2$ is directly trained on code completion.
It is because the syntax structures of the code are highly similar, which makes it easier for the model to learn.
The loss can not accurately reflect the discrepancies in operand values.
Additionally, representing these operand values with decimals in text format is not token-efficient, as they often require multiple tokens for representation.
Therefore, we first quantize the spatial position values 
to 8 bits, resulting in a maximum value of 256 (see Appendix \ref{app_quan}).
We apply a re-weighting loss $\mathcal{L}_{sgo}$ by doubling the loss values for numerical tokens to train the LLM of $\phi_2$:
\begin{equation}
\mathcal{L}_{sgo} = w_i \cdot \mathcal{L}_i,\ w_i = \begin{cases} 
2, & \text{if } y_i \in \mathbb{R}, \\
1, & \text{otherwise},
\end{cases}
\end{equation}
where $\mathcal{L}_{i}$ and $w_i$ denote loss and weight of the $i$-th token. $\mathbb{R}$ represents the set of real numbers.

\subsubsection{SFT for Error Correction}
The code editor $\phi_2$ takes the reference image, CAD program, rendered image, and generated feedback as inputs. Given that the vision encoder of $\phi_2$ is already sufficiently robust in recognizing geometric components, we only employ LoRA fine-tuning on the LLM for program error correction.

\begin{table*}[]
\centering
\begin{spacing}{1.1}
\resizebox{1.\columnwidth}{!}{
\caption{\label{tab_main} Main results of baselines and our ReCAD. $\dag$: fine-tuning on \textit{CADReview} dataset. CD, MMD and JSD are multiplied by 10$^3$. \textbf{Bold}: best results. }
\begin{tabular}{c|ccccccc|ccccccc}
\toprule[1pt]
\multirow{2}{*}{\textbf{Method}} & \multicolumn{7}{c|}{\textbf{Machine-made Program}}                                                                                                                                         & \multicolumn{7}{c}{\textbf{Human-made Program}}                                                                                                                                            \\ \cmidrule{2-15}
                         & \multicolumn{1}{c|}{\textbf{R$_L$}$\uparrow$} & \multicolumn{1}{c|}{\textbf{BS}$\uparrow$}    & \multicolumn{1}{c|}{\textbf{Acc}$\uparrow$}   & \multicolumn{1}{c|}{\textbf{CD}$\downarrow$}  & \multicolumn{1}{c|}{\textbf{MMD}$\downarrow$}   & \multicolumn{1}{c|}{\textbf{JSD}$\downarrow$}   & \textbf{IR}$\downarrow$    & \multicolumn{1}{c|}{\textbf{R$_L$}$\uparrow$} & \multicolumn{1}{c|}{\textbf{BS}$\uparrow$}    & \multicolumn{1}{c|}{\textbf{Acc}$\uparrow$}   & \multicolumn{1}{c|}{\textbf{CD}$\downarrow$}  & \multicolumn{1}{c|}{\textbf{MMD}$\downarrow$}   & \multicolumn{1}{c|}{\textbf{JSD}$\downarrow$}   & \textbf{IR}$\downarrow$    \\ \toprule
Claude 3.5                        & \multicolumn{1}{c|}{24.84}                                     & \multicolumn{1}{c|}{59.96}                                  & \multicolumn{1}{c|}{32.19}                                   & \multicolumn{1}{c|}{4.06}                & \multicolumn{1}{c|}{1.66}                                      & \multicolumn{1}{c|}{36.16}                                     & 32.97                                    & \multicolumn{1}{c|}{24.88}                                     & \multicolumn{1}{c|}{60.23}                                  & \multicolumn{1}{c|}{21.51}                                   & \multicolumn{1}{c|}{9.03}                                     & \multicolumn{1}{c|}{2.63}                                      & \multicolumn{1}{c|}{79.63}                                     & 86.09                                    \\ \hline
Gemini 2.0                        & \multicolumn{1}{c|}{27.08}                                     & \multicolumn{1}{c|}{63.29}                                  & \multicolumn{1}{c|}{36.83}                                   & \multicolumn{1}{c|}{3.96}                & \multicolumn{1}{c|}{0.45}                                      & \multicolumn{1}{c|}{11.49}                                     & 27.10                                    & \multicolumn{1}{c|}{25.61}                                     & \multicolumn{1}{c|}{59.95}                                  & \multicolumn{1}{c|}{23.36}                                   & \multicolumn{1}{c|}{5.97}                                     & \multicolumn{1}{c|}{1.60}                                      & \multicolumn{1}{c|}{55.85}                                     & 34.88                                    \\ \hline
GPT-4o                            & \multicolumn{1}{c|}{31.31}                                     & \multicolumn{1}{c|}{66.25}                                  & \multicolumn{1}{c|}{41.54}                                   & \multicolumn{1}{c|}{4.34}                & \multicolumn{1}{c|}{0.56}                                      & \multicolumn{1}{c|}{12.07}                                     & 28.43                                    & \multicolumn{1}{c|}{29.97}                                     & \multicolumn{1}{c|}{64.81}                                  & \multicolumn{1}{c|}{31.84}                                   & \multicolumn{1}{c|}{5.52}                                     & \multicolumn{1}{c|}{1.55}                                      & \multicolumn{1}{c|}{48.70}                                     & 18.57                                    \\ \hline
Llama 3.2$^\dag$                         & \multicolumn{1}{c|}{32.48}                                     & \multicolumn{1}{c|}{62.24}                                  & \multicolumn{1}{c|}{56.23}                                   & \multicolumn{1}{c|}{2.71}                & \multicolumn{1}{c|}{0.43}                                      & \multicolumn{1}{c|}{5.44}                                      & 4.01                                     & \multicolumn{1}{c|}{32.65}                                     & \multicolumn{1}{c|}{67.23}                                  & \multicolumn{1}{c|}{51.24}                                   & \multicolumn{1}{c|}{5.82}                                     & \multicolumn{1}{c|}{1.67}                                      & \multicolumn{1}{c|}{36.18}                                     & 16.14                                    \\ \hline
ReCAD-LA$^\dag$                          & \multicolumn{1}{c|}{\textbf{35.89}}                            & \multicolumn{1}{c|}{\textbf{70.24}}                         & \multicolumn{1}{c|}{\textbf{73.11}}                          & \multicolumn{1}{c|}{1.45}                & \multicolumn{1}{c|}{0.32}                                      & \multicolumn{1}{c|}{3.42}                                      & \textbf{0.00}                            & \multicolumn{1}{c|}{33.14}                                     & \multicolumn{1}{c|}{67.43}                                  & \multicolumn{1}{c|}{59.15}                                   & \multicolumn{1}{c|}{4.76}                                     & \multicolumn{1}{c|}{0.75}                                      & \multicolumn{1}{c|}{33.09}                                     & 13.74                                    \\ \hline
ReCAD-QW$^\dag$                          & \multicolumn{1}{c|}{35.62}                                     & \multicolumn{1}{c|}{70.01}                                  & \multicolumn{1}{c|}{71.83}                                   & \multicolumn{1}{c|}{\textbf{1.43}}       & \multicolumn{1}{c|}{\textbf{0.30}}                             & \multicolumn{1}{c|}{\textbf{2.80}}                             & \textbf{0.00}                            & \multicolumn{1}{c|}{\textbf{33.20}}                            & \multicolumn{1}{c|}{\textbf{67.47}}                         & \multicolumn{1}{c|}{\textbf{63.60}}                          & \multicolumn{1}{c|}{\textbf{4.42}}                            & \multicolumn{1}{c|}{\textbf{0.52}}                             & \multicolumn{1}{c|}{\textbf{30.87}}                            & \textbf{10.59}                           \\
 \midrule[1pt]
\end{tabular}
}
\end{spacing}
\end{table*}

\subsection{RL-based Refinement}
To ensure faithful feedback as guidance for code editing, we design two distinct reward functions to refine the feedback generator $\phi_1$. 
The first reward function leverages error diagnostics to judge the correctness of the generated feedback.
The second reward function is defined as the visual similarity between the rendered image of the edited code and the reference image. 
We further explore a time-consuming reward function based on 3D point clouds in Appendix \ref{app_reward}.

\vspace{0.5em}
\noindent
\textbf{Error Diagnostic.}  This diagnostic reward evaluates the correctness of identified code blocks and error types. 
The reward function $\mathcal{V}_d$ for two distinct states can be formulated as:
\begin{equation}
    \mathcal{V}_d=\left\{\begin{aligned}
1, & \text { if correct feedback, } \\
0, & \text { otherwise, }
\end{aligned}\right.
\end{equation}
where $\mathcal{V}_d$ equals 1 if both the code block and error type in the feedback are correctly identified.

\vspace{0.5em}
\noindent
\textbf{Visual Similarity.} 
We draw inspiration from self-similarity \cite{DeTikZify} and argue that the model can self-assess the similarity between the input reference image and the rendered output image. 
Thus, we obtain their visual features by the vision encoder of $\phi_1$ and calculate the cosine similarity of two features as the visual reward function $\mathcal{V}_v$.

\vspace{0.5em}
\noindent
\textbf{DPO.} 
We randomly choose 2,000 training samples and employ top-p sampling method to produce $K$ output feedback. Next, we collect them as rank pairs based on the two reward functions. 
We optimize $\phi_1$ with the direct preference optimization (DPO) \cite{dpo} algorithm. More details about the reward functions and refinement process are shown in Appendix \ref{app_reward}.

\begin{table}[]
\centering
\begin{spacing}{1.05}
\resizebox{1.\columnwidth}{!}{
\caption{\label{tab_abla} Results of ablation study. Fed: feedback. Red: reward functions. CD and JSD are multiplied by 10$^3$. \textbf{Bold}: best results.}
\begin{tabular}{c|c|ccc|ccc}
\toprule[1pt]
\multirow{2}{*}{\textbf{ReCAD}} & \multirow{2}{*}{\textbf{w/o}} & \multicolumn{3}{c|}{\textbf{Machine-made}}                                                  & \multicolumn{3}{c}{\textbf{Human-made}}                                                     \\ \cmidrule{3-8}
                       &                      & \multicolumn{1}{c|}{\textbf{Acc}$\uparrow$} & \multicolumn{1}{c|}{\textbf{CD}$\downarrow$} & \textbf{JSD}$\downarrow$ & \multicolumn{1}{c|}{\textbf{Acc}$\uparrow$} & \multicolumn{1}{c|}{\textbf{CD}$\downarrow$} & \textbf{JSD}$\downarrow$ \\ \toprule
\multirow{5}{*}{LA}             & GCR                           & \multicolumn{1}{c|}{67.43}                                   & \multicolumn{1}{c|}{2.08}                                     & 4.86                                      & \multicolumn{1}{c|}{51.75}                                   & \multicolumn{1}{c|}{5.86}                                     & 37.66                                     \\ \cline{2-8} 
                                & SGO                           & \multicolumn{1}{c|}{70.55}                                       & \multicolumn{1}{c|}{1.93}                                     & 4.03                                      & \multicolumn{1}{c|}{58.79}                                       & \multicolumn{1}{c|}{6.84}                                     & 39.24                                     \\ \cline{2-8} 
                                & Fed                           & \multicolumn{1}{c|}{-}                                       & \multicolumn{1}{c|}{1.84}                                     & 4.41                                      & \multicolumn{1}{c|}{-}                                       & \multicolumn{1}{c|}{5.71}                                     & 35.40                                     \\ \cline{2-8} 
                                & Red                           & \multicolumn{1}{c|}{71.16}                                   & \multicolumn{1}{c|}{1.57}                                     & 3.58                                      & \multicolumn{1}{c|}{58.01}                                   & \multicolumn{1}{c|}{5.27}                                     & 33.58                                     \\ \cline{2-8} 
                                & -                             & \multicolumn{1}{c|}{\textbf{73.11}}                          & \multicolumn{1}{c|}{\textbf{1.45}}                            & \textbf{3.42}                             & \multicolumn{1}{c|}{\textbf{59.15}}                          & \multicolumn{1}{c|}{\textbf{4.76}}                            & \textbf{33.09}                            \\ \midrule
\multirow{5}{*}{QW}             & GCR                           & \multicolumn{1}{c|}{66.28}                                   & \multicolumn{1}{c|}{2.03}                                     & 4.41                                      & \multicolumn{1}{c|}{53.74}                                   & \multicolumn{1}{c|}{6.03}                                     & 36.05                                     \\ \cline{2-8} 
                                & SGO                           & \multicolumn{1}{c|}{70.22}                                       & \multicolumn{1}{c|}{1.95}                                     & 4.27                                      & \multicolumn{1}{c|}{60.82}                                       & \multicolumn{1}{c|}{8.08}                                     & 42.56                                     \\ \cline{2-8} 
                                & Fed                           & \multicolumn{1}{c|}{-}                                       & \multicolumn{1}{c|}{1.59}                                     & 3.62                                      & \multicolumn{1}{c|}{-}                                       & \multicolumn{1}{c|}{5.49}                                     & 32.30                                     \\ \cline{2-8} 
                                & Red                           & \multicolumn{1}{c|}{71.26}                                   & \multicolumn{1}{c|}{1.80}                                     & 2.95                                      & \multicolumn{1}{c|}{61.79}                                   & \multicolumn{1}{c|}{4.74}                                     & 31.96                                     \\ \cline{2-8} 
                                & -                             & \multicolumn{1}{c|}{\textbf{71.83}}                          & \multicolumn{1}{c|}{\textbf{1.43}}                            & \textbf{2.80}                             & \multicolumn{1}{c|}{\textbf{63.60}}                          & \multicolumn{1}{c|}{\textbf{4.42}}                            & \textbf{30.87}                          
           \\ 
                       \midrule[1pt]
\end{tabular}
}
\end{spacing}
\end{table}

\section{Experiment}
\subsection{Experimental Setting}
\noindent
\textbf{Implementation Details.}
All experiments are conducted with four NVIDIA A100-80GB GPUs. 
We implement the ReCAD using different multimodal large language models (MLLMs), \ie, QWen2-VL ~\cite{qwen2-vl} and LLaVA-OV \cite{llava-ov}, each with approximately 7 billion parameters. These variants of ReCAD are named as ReCAD-QW and ReCAD-LA, respectively.
For the feedback generator $\phi_1$, we first only train the projector for 1 epoch on CAD caption. 
Next, we keep the projector frozen and fully fine-tune the large language model (LLM) and vision encoder for 1 epoch on CAD localization.
For feedback generation, we then apply learnable LoRA layers into the LLM of $\phi_1$, setting the rank to 8, a learning rate of 1e-5, 3 training epochs and a maximum sequence length of 3072.
For code editor $\phi_2$, we initialize it using the one trained on CAD caption and localization.
We train the LLM of $\phi_2$ on spatial geometric operation for 1 epoch.
Based on the generated feedback from $\phi_1$, we subsequently perform LoRA fine-tuning on $\phi_2$ to edit erroneous code, with the rank of 64. The batch size is 4 and we train 3 epochs with a learning rate of 4e-5. 
Moreover, the maximum sequence length of $\phi_2$ is expanded to 8192.
During the RL-based refinement, we use a top-p sampling strategy on $\phi_1$ with a temperature of 0.8 and $p$=0.9, generating 8 feedback samples for each input. 
Finally, we maintain those feedback pairs where the diagnostic reward of chosen feedback equals 1, and the visual reward difference exceeds 0.25 for direct preference optimization.

\vspace{0.5em}
\noindent
\textbf{Evaluation Metrics.}
Similar to previous studies on code review \cite{coffee,codeagent},   we employ metrics on feedback generation and program error correction to comprehensively evaluate the performance of our ReCAD and baselines.
Following \cite{causal-vqg, l2c}, we adopt natural language generation metrics, \ie, ROUGE$_L$ (R$_L$) ~\cite{rouge}, BERTScore (BS) ~\cite{bert-score}, and accuracy (Acc) to assess the quality of generated feedback. 
Notably, accuracy is a rigorous criterion, requiring both the code block and error type to be correctly predicted.
Following \cite{deepcad}, we utilize chamfer distance (CD), minimum matching distance (MMD), Jensen-Shannon divergence (JSD), and invalidity ratio (IR) to evaluate the consistency between 3D point clouds of the edited and ground truth 3D objects. We detail these metrics in Appendix \ref{app_metric}.

\vspace{0.5em}
\noindent
\textbf{Baselines.}
To verify the superiority of our ReCAD, we compare with several advanced closed-source multimodal large language models (MLLMs), \ie, GPT-4o \cite{gpt4o}, Claude 3.5 \cite{claude3-5}, and Gemini 2.0 \cite{gemini2}.
Moreover, we fine-tune a larger open-source MLLM, \ie, Llama-3.2-11B \cite{llama-3} for additional comparison.
We also report the results of baselines that directly edit code without feedback and apply the few-shot setting in Appendix \ref{app_res}.

\subsection{Performance Comparison}
Table \ref{tab_main} reports the quantitative evaluation results of baselines and our ReCAD framework on \textit{CADReview} dataset. 
We have the following observations: 
\textbf{(i)} While the leading closed-source MLLM, GPT-4o, remains competitive on natural language generation metrics (\ie, ROUGE$_L$ and BERTScore), it only achieves 41.54\% and 31.84\% accuracy in generating feedback on human-made and machine-made programs, respectively, lagging behind our ReCAD by over 30\%.
It suggests that they exhibit similar capabilities in describing visual anomalies, 
but close-source MLLMs struggle to align CAD programs with visual information for feedback generation, leading to poor performance in identifying errors within CAD programs.
Moreover, they often generate non-compilable CAD programs, with an invalid rate exceeding 25\%.
\textbf{(ii)} Our ReCAD-LA and ReCAD-QW consistently outperform the larger open-source Llama 3.2 on all metrics, even though these three models have been fine-tuned on the \textit{CADReview} dataset. 
For example, on human-made programs, our ReCAD-QW surpasses Llama 3.2 by ``+12.36'' and ``-5.31'' on Acc and JSD, respectively.
This gap highlights that fine-tuning alone is insufficient for the CAD review task. Our geometric component recognition (GCR) and spatial geometric operation (SGO) mechanisms assist MLLMs in generating accurate feedback and translating geometric operations into precise code implementations. 
\textbf{(iii)} 
Compared to the machine-made program, all methods experience a performance drop on the human-made program. 
Our ReCAD still maintains state-of-the-art performance on the human-made program.
It demonstrates that ReCAD not only learns geometric operations for simple primitives in machine-made programs but also handles complex geometric components and program structures in human-made programs, resulting in more accurate reconstructed 3D objects.
\textbf{(iv)} Despite ReCAD-LA achieving higher accuracy in feedback generation on machine-made programs, ReCAD-QW consistently leads in the quality of reconstructed 3D objects on two types of programs. 
It shows that the feedback generated by feedback generator $\phi_1$ of ReCAD-QW provides more effective guidance for code editor $\phi_2$ to correct errors.

\subsection{Ablation Study}
Table \ref{tab_abla} shows ablation experimental results. 
We find that:
\textbf{(i)} Removing the geometric component recognition (GCR) mechanism degrades ReCAD's performance on all metrics. 
The feedback generator $\phi_1$ fails to align code blocks with their corresponding components for feedback generation. This misalignment also adversely affects the initialization of code editor $\phi_2$, preventing accurate program error correction. 
As a result, the quality of the reconstructed 3D objects is even inferior to the variant where no feedback is provided to $\phi_2$ (\ie, ReCAD w/o Fed).
\textbf{(ii)} 
The spatial geometric operation (SGO) mechanism is crucial in handling human-made programs, which often feature more complex geometric components (\eg, hidden internal components) and program structures. 
It validates that $\phi_2$ effectively captures these factors through the SGO mechanism to further edit the erroneous code. 
\textbf{(iii)} By comparing ReCAD w/o Fed, ReCAD w/o Red and ReCAD, we observe that accurate feedback significantly enhances the ability of $\phi_2$ to edit code. 
Furthermore, we notice a 
slight performance improvement by applying the reward functions $\mathcal{V}_d$ and $\mathcal{V}_v$.
They evaluate the quality of both the feedback and the edited code to refine $\phi_1$, 
which ensure cycle consistency between $\phi_1$ and $\phi_2$.
\subsection{Case Study}
\begin{figure}[!]
  \centering
  \includegraphics[scale=0.471]{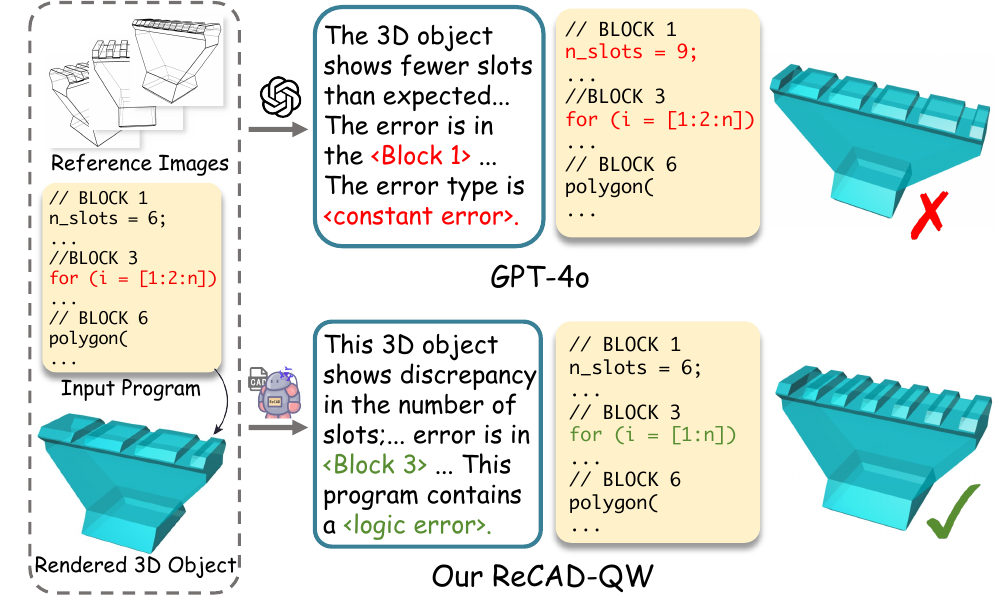}
  \caption{Cases of different methods. The \textcolor[RGB]{88,142,49}{green} and \textcolor[RGB]{255,0,0}{red} denote correct and incorrect prediction, respectively.}
\label{pic_case_study}
\end{figure}
Fig. \ref{pic_case_study} shows the generated feedback, edited code and corresponding reconstructed 3D object by GPT-4o and our ReCAD-QW.
Intuitively, our ReCAD-QW produces a more grounded 3D object. 
Specifically, although both GPT-4o and ReCAD-QW correctly identify the visual discrepancies regarding the number of slots in the 3D object, our ReCAD-QW successfully locates the erroneous loop statement in block 6 for program error correction.
In contrast, GPT-4o relies solely on the semantics of the variable name, which results in the edited code containing even more errors.

\section{Conclusion}
In this paper, we introduce the CAD review task to automatically detect and correct errors in CAD programs based on reference images. 
To tackle this task, we propose the \textit{CADReview} dataset and ReCAD framework to detect program errors and generate helpful feedback on error correction. Extensive experiments show that ReCAD outperforms existing MLLMs, providing an effective solution for the AI-aided review process in industrial design.

\section{Limitations}
In this paper, we introduce the CAD review task. We also create the \textit{CADReview} dataset and propose the ReCAD framework for this task.
Although our framework aids designers in the design review process, it does not consider the optimal code editing solution. Given the diversity in programmatic realizations of the same geometric components, the optimal solution is hard to define. 
Additionally, due to limitations about copyright and data availability of CAD programming languages, our \textit{CADReview} dataset currently only includes open-source OpenSCAD code.
Since all CAD programming languages adhere to common standards (\eg, basic geometric primitives and operations), we believe the ReCAD framework is language-agnostic and can be adapted to any CAD programming language in the future.

\section*{Acknowledgments}
This research is supported by Guangdong Provincial Natural Science Foundation for Outstanding Youth Team Project (2024B1515040010),  the Fundamental Research Funds for the Central Universities, South China University of Technology (x2rjD2240100),  Guangdong Provincial Fund for Basic and Applied Basic Research—Regional Joint Fund Project (Key Project) (2023B1515120078).

\bibliography{custom.bib}

\newpage
\appendix
\section{\textit{CADReview} Dataset Construction} \label{app_dataset}
\begin{figure*}[]
  \centering
  \includegraphics[scale=0.238]{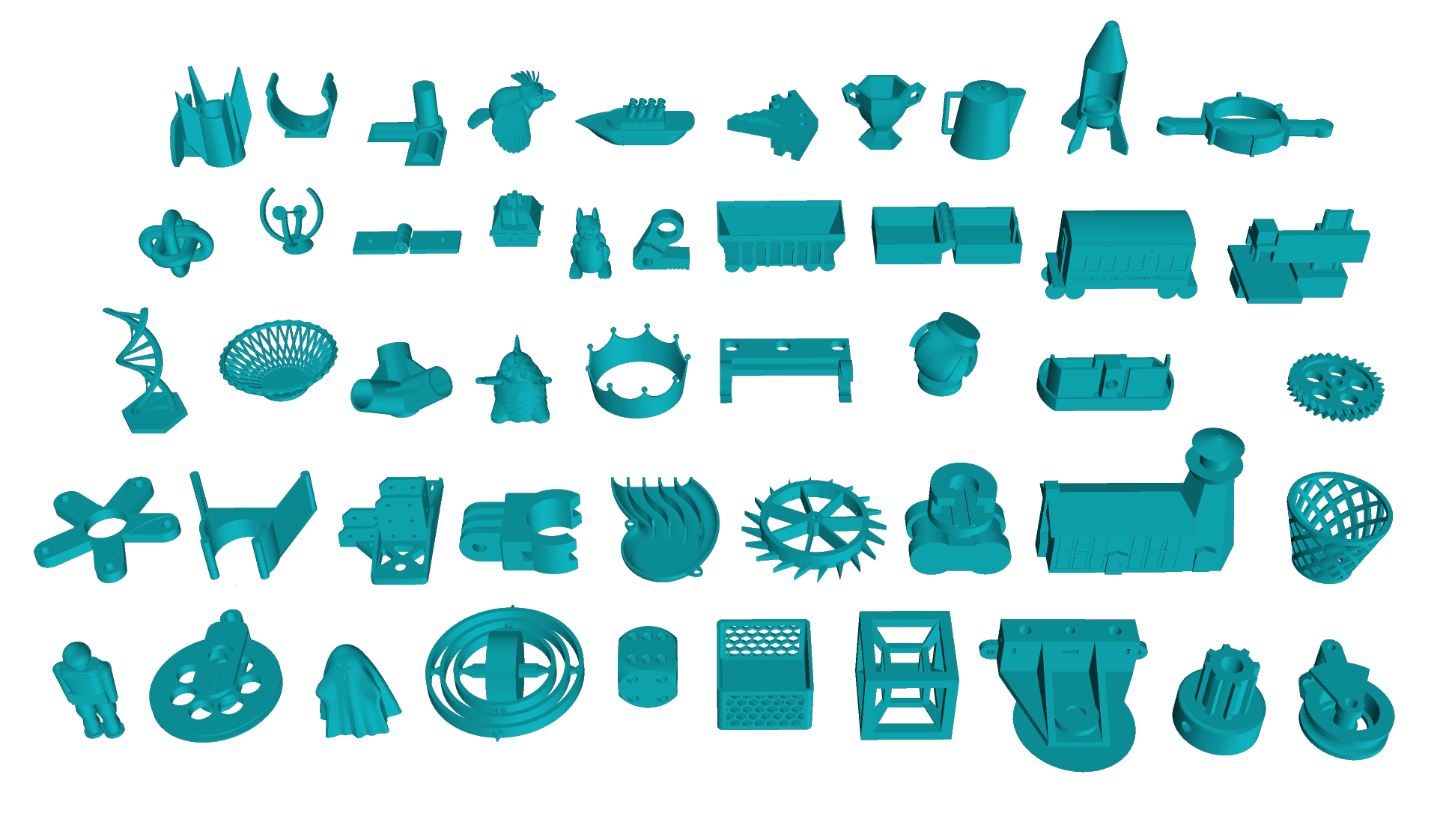}
  \caption{Demonstration of various 3D objects from our \textit{CADReview} dataset.
  } 
  \label{pic_data_demo}
\end{figure*}

In this section, we provide additional details about the construction of our \textit{CADReview} dataset.
Fig. \ref{pic_data_demo} shows the demonstration of diverse 3D objects in our \textit{CADReview} dataset.

\subsection{CAD Program Collection} \label{app_data_overview}
\begin{figure}[]
  \centering
  \includegraphics[scale=0.35]{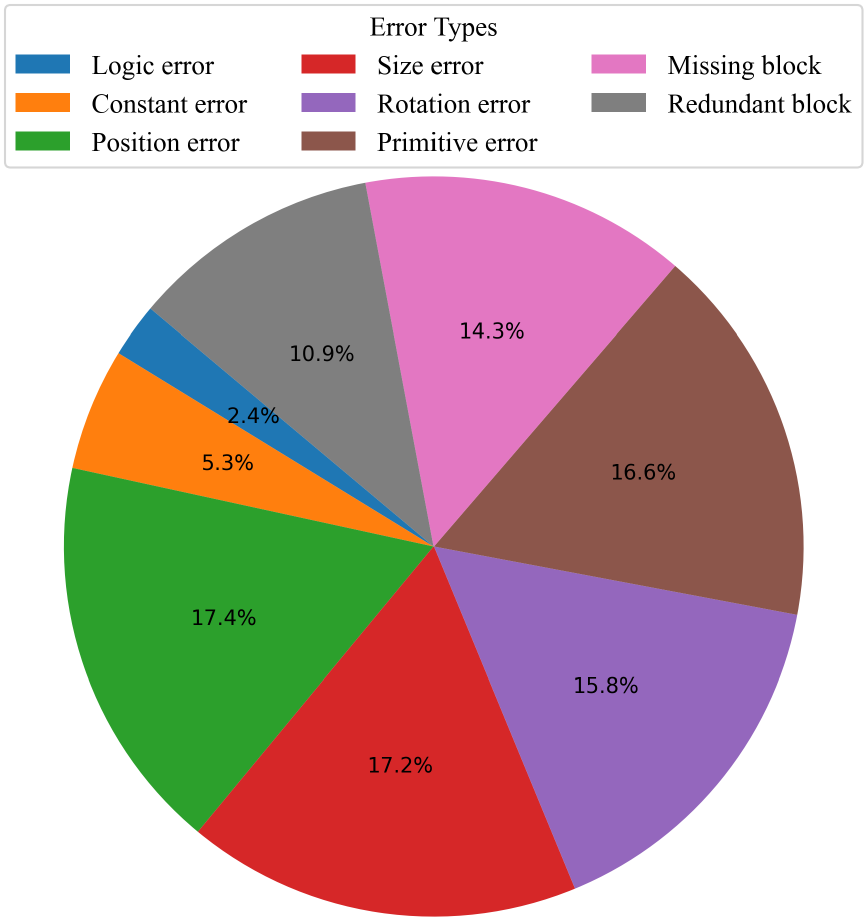}
  \caption{Distribution statistics of error types in our \textit{CADReview} dataset.
  } 
  \label{pic_error_sta}
\end{figure}

\begin{figure*}[]
  \centering
  \includegraphics[scale=0.61]{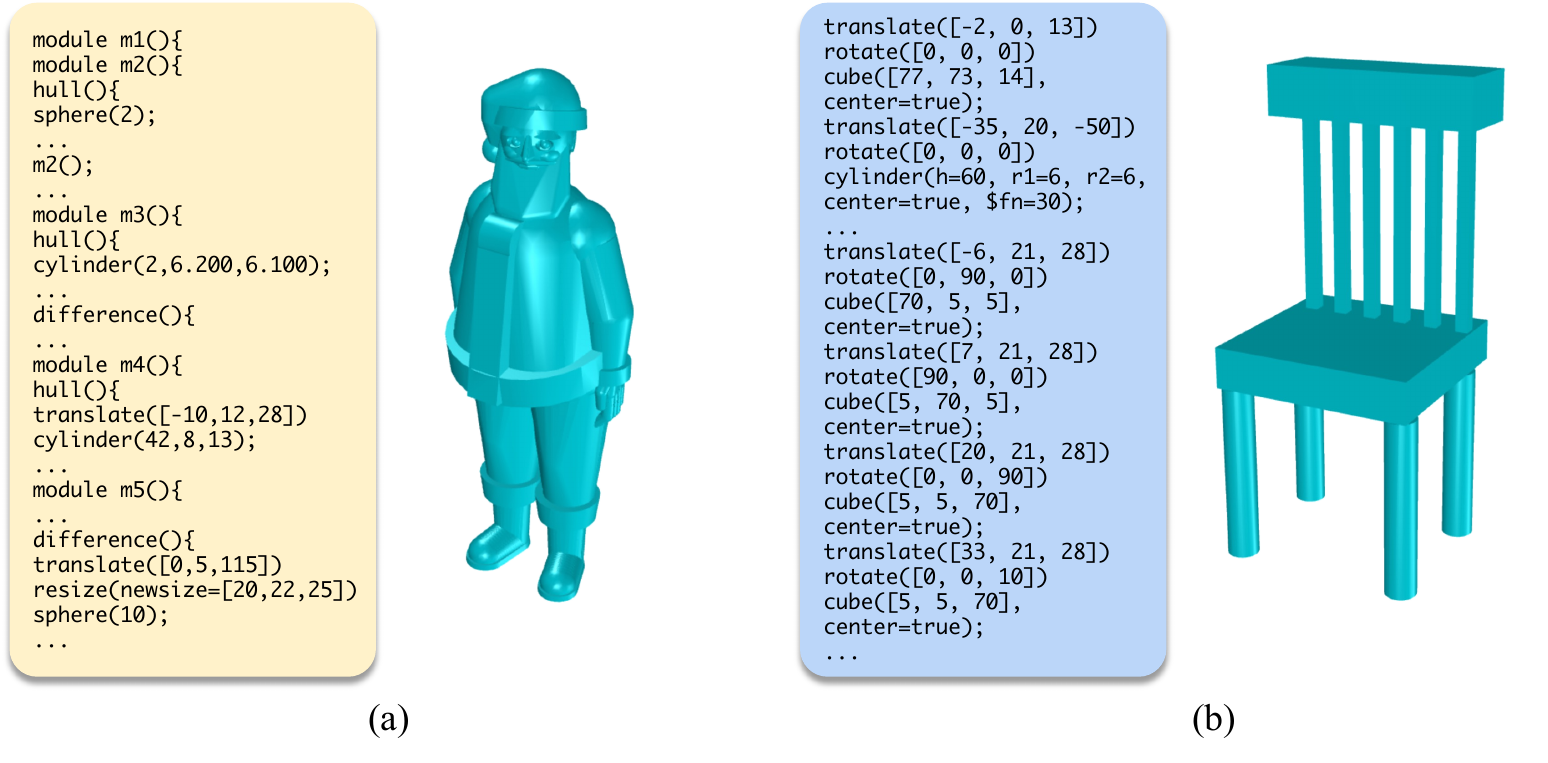}
  \caption{Samples from \textit{CADReview} dataset. (a) Human-made program. (b) Machine-made program.
  } 
  \label{pic_pro}
\end{figure*}

The source data of the human-made programs are collected from online design communities\footnotemark\footnotetext{https://cults3d.com}. 
We manually remove comments and redundant code segments to simplify the program structure.
Our dataset consists of 21,949 samples, including 9,218 human-made programs and 12,731 machine-generated programs. 
It is worth noting that each set (\ie, training or testing set) contains CAD programs from a set of mutually exclusive 3D objects. 
Therefore, the evaluation is conducted on previously unseen 3D objects, which allows for a better assessment of the model's generalization ability.
Fig. \ref{pic_pro} illustrates samples of human-made and machine-made programs in our dataset.
The distribution of error types is presented in Fig. \ref{pic_error_sta}.

\subsection{Reference Image Collection}
We utilize 3D modeling software (\ie, Rhino) and adopt pen-style mode for rendering to obtain the reference images.
Specifically, for these 3D objects, we also employ a perspective view with the pen-style mode for rendering, allowing the potential internal components to be visible. 
Each viewpoint of the reference image is selected from the range of $[0, 2\pi)$, with the condition that the angular difference between consecutive viewpoints must be greater than $\pi/10$. 
Once the reference images are selected, they are provided to another annotator to evaluate whether these images sufficiently capture the essential structure of the 3D object.

\subsection{Error Creation} \label{app_error_create}
\begin{figure*}[]
  \centering
  \includegraphics[scale=0.118]{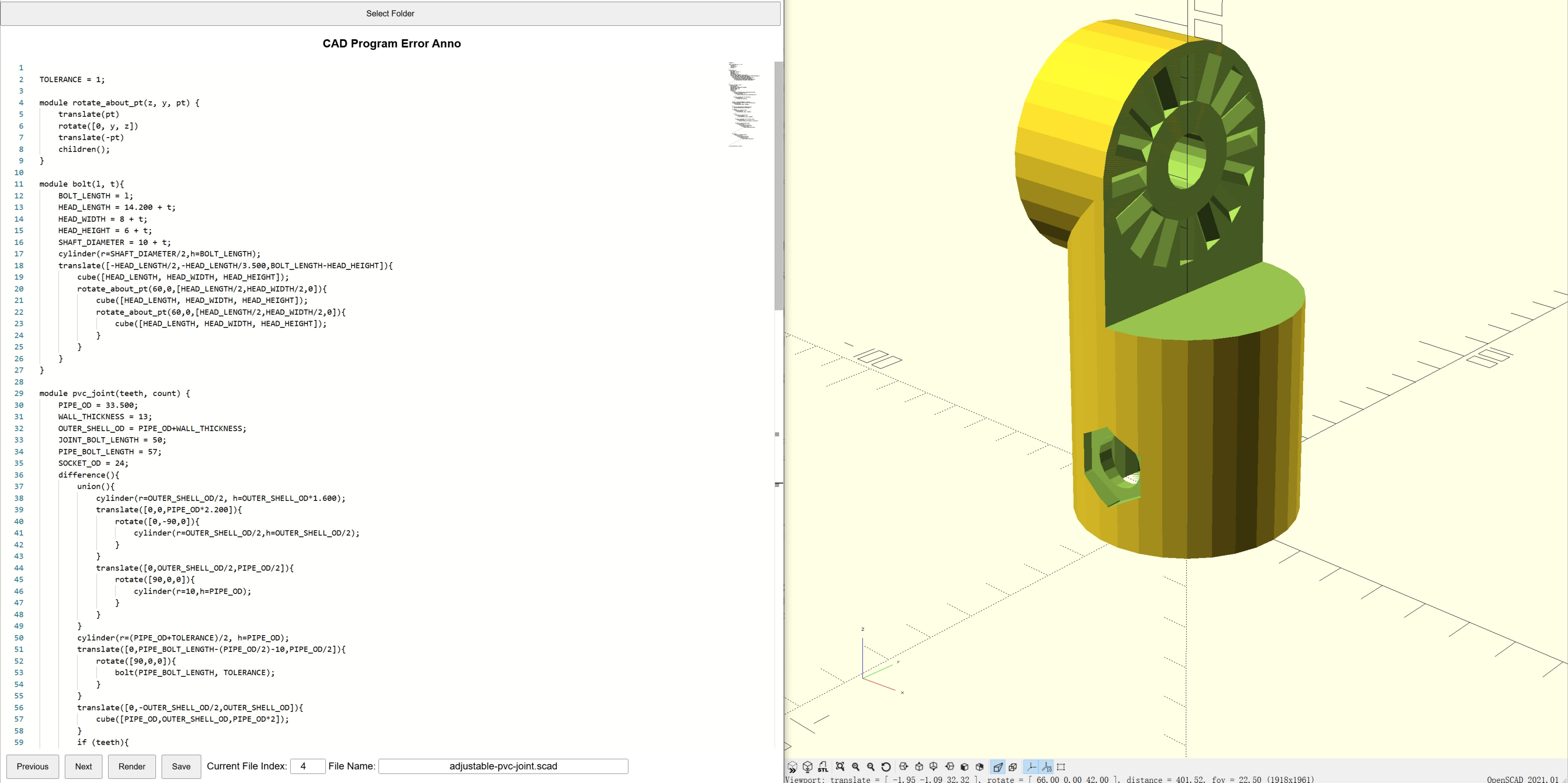}
  \caption{Graphic annotation interface for error creation.
  } 
  \label{pic_interface}
\end{figure*}

We develop a graphical annotation interface for annotators to create errors on the collected CAD programs, as shown in Fig. \ref{pic_interface}.
Our annotation team consists of 10 experienced designers familiar with OpenSCAD code for error annotation.

\subsection{Feedback Annotation} \label{app_feedback}
We provide the prompt for GPT-4o to annotate feedback in Table \ref{tab_feedback_prompt}.
Furthermore, for CAD programs that are consistent with the reference images (\ie, correct programs), we randomly select one of the 10 predefined feedback options listed in Table \ref{tab_cor_feedback}.

\section{Additional Training Dataset} \label{app_pretrain}
To construct training data for CAD caption and CAD localization, we utilize the existing Text2CAD \cite{text2cad} and CADTalk ~\cite{cadtalk} datasets. 
Considering that 3D objects from realistic industrial design often feature hidden internal components, we further augment data by GPT-4o.

\subsection{Text2CAD}
Text2CAD dataset is built on the original DeepCAD \cite{deepcad} dataset. 
It leverages large language models (LLMs) and vision-language models (VLMs) to generate text prompts, enabling the translation from textual descriptions to parametric CAD models. 
Initially, LLaVA-NeXT \cite{llava-next} is employed to produce shape descriptions of these CAD models and their intermediate components. 
Subsequently, Mixtral-50B \cite{mixtral} generates multi-level textual modeling instructions based on the shape descriptions and design details provided in DeepCAD. 
We utilize beginner-level instructions that describe how CAD models are composed of simple components. Meanwhile, for each instruction, we select four different views of the corresponding image as training data, effectively multiplying the total number of training samples by four. It results in over 600\textit{K} training samples for CAD caption.
Moreover, we prompt the MLLM to describe the composition of the 3D object during training for CAD caption.

\subsection{CADTalk}
CADTalk \cite{cadtalk} is a benchmark dataset for semantic comment annotation of CAD programs, where each code block is annotated to represent the semantic components of 3D objects. For example, Block 4 represents the hand of the 3D object.
It contains approximately 6.5\textit{K} samples, most of which are derived from labeled 3D objects in the PartNet \cite{partnet} dataset and converted into OpenSCAD programs. 
Another portion is collected from online repositories (\eg., Thingiverse\footnotemark\footnotetext{https://www.thingiverse.com}.
We leverage CADTalk to ground the coordinates of each semantic label in rendered images for semantic matching in CAD localization. 
For coordinate matching and localization, we randomly combine code blocks from the same sample program in CADTalk, resulting in over 10K data samples.

\subsection{LLM-Augmented Data} \label{app_llm_aug}
Considering that real 3D objects often feature complex internal components, we further synthesize LLM-augmented data using boolean operations (\eg, difference and union) with GPT-4. Subsequently, we design rules to randomly modify the values of the generated data for data augmentation. Finally, we obtain over 10\textit{K} data samples.
In particular, we collect two types of rendered images using both the pen-style and physical rendering modes with a perspective view, which enhances the model's generalization ability.
This LLM-augmented data is used for coordinate matching and localization in CAD localization.
As shown in Fig. \ref{pic_cad_loc} (b), these augmented programs consist of basic primitives and boolean operations without representation of tangible objects. 
The prompt for GPT-4o is provided in Table \ref{llm_aug_data_anno}.

\subsection{Value Quantization} \label{app_quan}
Considering that CADTalk and LLM-augmented data often contain many decimal values, MLLMs predict continuous parameters through regression, and slight inaccuracies can easily violate these critical constraints.
As a result, it is challenging for basic MLLMs to understand spatial relationships within CAD programs for spatial geometric operations (SGO). Following previous studies \cite{cad-signet,text2cad}, we quantize the spatial position values (\ie, axis coordinates and translation offset values) of the CAD programs to 8 bits, yielding a maximum value of 256.

\subsection{Performance Evaluation}
After introducing the training data, we briefly present the performance of our ReCAD on these tasks, which demonstrates the effectiveness of our geometric component recognition (GCR) and spatial geometric operations (SGO) mechanisms.
Specifically, for CAD caption, CAD localization, and code completion, we sample 500 instances from each task for performance evaluation.

For CAD caption, we observe a significant improvement in performance on natural language generation metrics (\eg, BLEU-4 \cite{bleu} and BERTScore \cite{bert-score}). For instance, after training the model on CAD caption, the BLEU-4 score increases from 4.42 to 20.03.
It suggests that the projector effectively refines visual features, enabling our ReCAD to focus on geometric components.
For CAD localization, we set the threshold of Intersection over Union (IoU) \cite{iou} between the predicted and ground-truth bounding boxes to 0.8, considering predictions with an IoU below this value as failures.
The overall accuracy of our ReCAD-LA and ReCAD-QW are 91.06\% and 88.25\%, respectively.
After sequential training on CAD caption and localization, we notice that the performance of CAD caption remains unaffected.
It demonstrates that our ReCAD framework effectively aligns identified geometric components with their corresponding code blocks.
Finally, for code completion in spatial geometric operations (SGO), the trained model achieves a chamfer distance of less than 1 between the 3D object generated from the completed code and the ground truth, whereas the initial model is unable to generate compilable code. 
It suggests that the SGO mechanism enables our ReCAD framework to effectively learn spatial operations of basic primitives and facilitates further code editing in the CAD review task.

\section{Additional Results} \label{app_res}

\subsection{Human Evaluation}
Considering that the same geometric shape can be realized through different programmatic implementations, we evaluate the code editing solutions of our ReCAD and the baselines by comparing them to those of human designers with human evaluation.
We first randomly select 200 samples with a chamfer distance smaller than 5 in human-made programs from our ReCAD and Llama 3.2 prediction results, and both predicted code blocks and error types are correct. 
The selected samples are consistent across all methods.
We invite five volunteer designers familiar with OpenSCAD to evaluate  
based on the following criteria: Readability (\textbf{Red}) measures how clear and understandable is the edited code to a human designer. Reusability (\textbf{Reu}) refers to what extent can the edited code be reused or extended to other 3D objects.
Preservation \textbf{(Pre)} represents how well the edited code maintains the original functionality while eliminating errors.
The scores on a scale from 0 to 2, with higher values indicating greater alignment with human-edited code. 
Table \ref{tab_human} shows the results of our ReCAD and Llama 3.2. The standard deviations of each human evaluation metric confirms the faithfulness of our results.
We find that: \textbf{(i)} After fine-tuning, these methods demonstrate consistent performance in generating readable code, suggesting that they have effectively learned the syntax rules of CAD programs.
\textbf{(ii)} Our ReCAD significantly outperforms the Llama 3.2 model in terms of code reusability and functionality preservation, demonstrating that our mechanisms for geometric component recognition (GCR) and spatial geometric operations (SGO) enable more adaptable CAD program generation.

\subsection{Few-Shot Setting for Baselines}
In Table \ref{tab_few_shot}, we report the few-shot results for close-source MLLM baselines (\ie, Claude 3.5, Gemini 2.0 and GPT-4o).
Specifically, we randomly select 200 samples in machine-made and human-made programs, respectively.
We utilize CodeBERT \cite{codebert} to encode CAD programs and compute the cosine similarity between the query program and each sample in the training set to find the most similar one as the 1-shot in-context example.
We find that the fluctuations in the results are minimal, suggesting that the few-shot approach does not effectively address the challenges of the CAD review task.

\begin{table}[]
\centering
\caption{\label{tab_human}Human evaluation results. Each value is presented as $\tau/\rho$, where $\tau$ is the metric value and $\rho$ is the standard deviation. 
\textbf{Bold}: the maximum value. }
\begin{spacing}{1.}
\resizebox{1.\columnwidth}{!}{
\begin{tabular}{l|c|c|c}
\toprule[1pt]
\textbf{Method} & \textbf{Red} & \textbf{Reu} & \textbf{Pre} \\ \hline
Llama 3.2       & 1.36/0.21    & 0.94/0.27    & 0.87/0.26    \\ \hline
ReCAD-LA        & \textbf{1.38}/0.30    & 1.25/0.21    & 1.23/0.22    \\ \hline
ReCAD-LA        & 1.31/0.19    & \textbf{1.29}/0.18    & \textbf{1.28}/0.24 \\ \midrule[1pt]
\end{tabular}}
\end{spacing}
\end{table}

\begin{table}[]
\centering
\caption{\label{tab_few_shot} Few-shot results for close-source MLLMs. CD is multiplied by 10$^3$.
\textbf{Bold}: the maximum value. }
\begin{spacing}{1.}
\resizebox{1.\columnwidth}{!}{
\begin{tabular}{c|c|cc|cc}
\toprule[1pt]
\multirow{2}{*}{\textbf{Model}}      & \multirow{2}{*}{\textbf{Few-shot}} & \multicolumn{2}{c|}{\textbf{Machine-made}} & \multicolumn{2}{c}{\textbf{Human-made}}    \\ \cmidrule{3-6} 
                            &                          & \multicolumn{1}{c|}{\textbf{Acc}$\uparrow$}   & \textbf{CD$\downarrow$}   & \multicolumn{1}{c|}{\textbf{Acc}$\uparrow$}   & \textbf{CD}$\downarrow$   \\ \toprule
\multirow{2}{*}{Claude 3.5} & \ding{55}                        & \multicolumn{1}{c|}{35.00} & \textbf{3.94} & \multicolumn{1}{c|}{\textbf{24.00}} & \textbf{9.76} \\  
                            & \ding{51}                       & \multicolumn{1}{c|}{\textbf{36.00}} & 3.90 & \multicolumn{1}{c|}{22.50} & 9.73 \\ \midrule
\multirow{2}{*}{Gemini 2.0} & \ding{55}                        & \multicolumn{1}{c|}{38.50} & 3.57 & \multicolumn{1}{c|}{\textbf{26.00}} & \textbf{6.01} \\  
                            & \ding{51}                       & \multicolumn{1}{c|}{\textbf{40.50}} & \textbf{3.81} & \multicolumn{1}{c|}{\textbf{26.00}} & 5.99 \\ \midrule
\multirow{2}{*}{GPT-4o}     & \ding{55}                        & \multicolumn{1}{c|}{43.50} & \textbf{4.16} & \multicolumn{1}{c|}{34.00} & 5.58 \\  
                            & \ding{51}                       & \multicolumn{1}{c|}{\textbf{45.00}} & 4.09 & \multicolumn{1}{c|}{\textbf{35.00}} & \textbf{5.70} \\ \midrule[1pt]
\end{tabular}}
\end{spacing}
\end{table}

\subsection{Direct Code Editing for Baselines}
Table \ref{tab_dir_edit} shows the results for close-source MLLMs to directly editing CAD code without feedback generation.
 Interestingly, the absence of feedback leads to performance improvements for both GPT-4o and Gemini 2.0. This contrasts with the findings from our ReCAD ablation study in Table \ref{tab_abla}, where the lack of feedback results in a significant performance decline for ReCAD.
This suggests that closed-source MLLMs are unable to effectively utilize feedback to enhance their ability to edit CAD programs, which is in contrast to code review tasks  \cite{coffee} on commonly used programming languages such as C++ and Python.
\begin{table}[]
\centering
\caption{\label{tab_dir_edit} Direct code editing results for close-source MLLMs. CD and JSD are multiplied by 10$^3$.
\textbf{Bold}: the maximum value. }
\begin{spacing}{1.}
\resizebox{1.\columnwidth}{!}{
\begin{tabular}{c|c|cc|cc}
\toprule[1pt]
\multirow{2}{*}{\textbf{Model}}      & \multirow{2}{*}{\textbf{Feedback}} & \multicolumn{2}{c|}{\textbf{Machine-made}}                   & \multicolumn{2}{c}{\textbf{Human-made}}                      \\ \cmidrule{3-6} 
                            &                           & \multicolumn{1}{c|}{\textbf{CD} $\downarrow$}            & \textbf{JSD$\downarrow$}            & \multicolumn{1}{c|}{\textbf{CD}$\downarrow$}            & \textbf{JSD$\downarrow$}            \\ \toprule
\multirow{2}{*}{Claude 3.5} & \ding{55}                       & \multicolumn{1}{c|}{4.46}          & 36.75          & \multicolumn{1}{c|}{9.05}          & 82.18          \\
                            & \ding{51}                       & \multicolumn{1}{c|}{\textbf{4.06}} & \textbf{36.16} & \multicolumn{1}{c|}{\textbf{9.03}} & \textbf{79.63} \\ \hline
\multirow{2}{*}{Gemini 2.0} & \ding{55}                       & \multicolumn{1}{c|}{\textbf{3.88}} & \textbf{9.54}  & \multicolumn{1}{c|}{\textbf{4.89}} & \textbf{53.75} \\
                            & \ding{51}                       & \multicolumn{1}{c|}{3.96}          & 11.49          & \multicolumn{1}{c|}{5.97}          & 55.85          \\ \hline
\multirow{2}{*}{GPT-4o}     & \ding{55}                       & \multicolumn{1}{c|}{\textbf{3.97}} & \textbf{8.94}  & \multicolumn{1}{c|}{\textbf{5.23}} & \textbf{45.94} \\
                            & \ding{51}                       & \multicolumn{1}{c|}{4.34}          & 12.07          & \multicolumn{1}{c|}{5.52}          & 48.70    
 \\ \midrule[1pt]
\end{tabular}}
\end{spacing}
\end{table}

\begin{table}[]
\centering
\begin{spacing}{1.}
\resizebox{1.\columnwidth}{!}{
\caption{\label{tab_reward} Results of different reward functions. CD and JSD are multiplied by 10$^3$. \textbf{Bold}: best results.}
\begin{tabular}{c|c|ccc|ccc}
\toprule[1pt]
\multirow{2}{*}{\textbf{ReCAD}} & \multirow{2}{*}{\textbf{Reward}} & \multicolumn{3}{c|}{\textbf{Machine-made}}                                                                                                                               & \multicolumn{3}{c}{\textbf{Human-made}}                                                                                                                                  \\ \cmidrule{3-8} 
                                &                                  & \multicolumn{1}{c|}{\textbf{Acc}$\uparrow$} & \multicolumn{1}{c|}{\textbf{CD}$\downarrow$} & \textbf{JSD}$\downarrow$ & \multicolumn{1}{c|}{\textbf{Acc}$\uparrow$} & \multicolumn{1}{c|}{\textbf{CD}$\downarrow$} & \textbf{JSD}$\downarrow$ \\ \toprule
\multirow{4}{*}{LA}             & -                                & \multicolumn{1}{c|}{71.16}                                   & \multicolumn{1}{c|}{1.57}                                     & 3.58                                      & \multicolumn{1}{c|}{58.01}                                   & \multicolumn{1}{c|}{5.27}                                     & 33.58                                     \\
                                & $\mathcal{V}_d$                  & \multicolumn{1}{c|}{72.49}                                   & \multicolumn{1}{c|}{1.52}                                     & 3.49                                      & \multicolumn{1}{c|}{59.00}                                   & \multicolumn{1}{c|}{4.85}                                     & 33.43                                     \\
                                & $\mathcal{V}_v$                  & \multicolumn{1}{c|}{71.89}                                   & \multicolumn{1}{c|}{1.58}                                     & 3.56                                      & \multicolumn{1}{c|}{58.39}                                   & \multicolumn{1}{c|}{4.91}                                     & 33.54                                     \\
                                & $\mathcal{V}_p$                  & \multicolumn{1}{c|}{\textbf{72.75}}                          & \multicolumn{1}{c|}{\textbf{1.46}}                            & \textbf{3.47}                             & \multicolumn{1}{c|}{\textbf{59.41}}                          & \multicolumn{1}{c|}{\textbf{4.68}}                            & \textbf{32.59}                            \\ \midrule
\multirow{4}{*}{QW}             & -                                & \multicolumn{1}{c|}{71.26}                                   & \multicolumn{1}{c|}{1.80}                                     & 2.95                                      & \multicolumn{1}{c|}{61.79}                                   & \multicolumn{1}{c|}{4.74}                                     & 31.96                                     \\
                                & $\mathcal{V}_d$                  & \multicolumn{1}{c|}{71.52}                                   & \multicolumn{1}{c|}{1.45}                                     & 2.84                                      & \multicolumn{1}{c|}{62.88}                                   & \multicolumn{1}{c|}{4.46}                                     & 31.11                                     \\
                                & $\mathcal{V}_v$                  & \multicolumn{1}{c|}{71.49}                                   & \multicolumn{1}{c|}{1.51}                                     & 2.91                                      & \multicolumn{1}{c|}{62.20}                                   & \multicolumn{1}{c|}{4.60}                                     & 31.87                                     \\
                                & $\mathcal{V}_p$                  & \multicolumn{1}{c|}{\textbf{72.09}}                          & \multicolumn{1}{c|}{\textbf{1.38}}                            & \textbf{2.78}                             & \multicolumn{1}{c|}{\textbf{63.16}}                          & \multicolumn{1}{c|}{\textbf{4.44}}                            & \textbf{31.18}                                         
           \\ 
                       \midrule[1pt]
\end{tabular}
}
\end{spacing}
\end{table}

\subsection{Influence of Rendered Image}
In our main experiment, we compile the input CAD programs and render an image as supplementary input information. This rendered image is easily obtained and feasible from an implementation perspective.
Moreover, our preliminary experiments show that omitting the rendered image results in nearly a 5\% drop in feedback accuracy for human-made programs. By analyzing the cases of prediction failure, we find that, without the rendered image, the model struggles to correctly identify erroneous code blocks in 3D objects with more than 10 geometric components.
It indicates that the rendered image also plays an important role in aligning components, which is consistent with our motivation of geometric component recognition.
Further experiments with multiple viewpoints of the rendered show minimal performance improvement for ReCAD, suggesting that a single rendered image is sufficient as supplementary input. 

\subsection{Reward Functions} \label{app_reward}
In our RL-based refinement process, we design three reward functions to refine the feedback generator $\phi_1$. 
The first reward function $\mathcal{V}_d$ leverages error diagnostics to judge the correctness of the generated feedback.
The second reward function $\mathcal{V}_v$ is defined as the visual similarity between the rendered image of the edited code and the reference image. 
The third reward function $\mathcal{V}_p$ is based on the sampling 3D point clouds.
During the RL-based refinement, we use a top-p sampling strategy on $\phi_1$ with a temperature of 0.8 and $p$=0.9, generating 8 feedback samples for each input. 
Table \ref{tab_reward} shows the comparison of different reward functions.
For $\mathcal{V}_d$ and $\mathcal{V}_v$, we maintain those feedback pairs for DPO, where the diagnostic reward of chosen feedback equals 1, or the visual reward difference exceeds 0.25.
For $\mathcal{V}_p$, we retain feedback pairs where the generated candidate 3D objects have the smallest chamfer distance to the ground truth, as the chosen samples. Conversely, the candidates with the largest chamfer distance to the ground truth are treated as reject samples. We observe that, although $\mathcal{V}_p$ yields the best performance improvement for our ReCAD, the process of sampling 3D point clouds and calculating distances is time-consuming. Therefore, we only present the results with $\mathcal{V}_d$ and $\mathcal{V}_v$ in the main experiment.

Moreover, consistent with the finding in code review for generic programming languages \cite{coffee}, we find that high-quality feedback is crucial for successful code editing. Therefore, we apply DPO solely to $\phi_1$. Our attempts to apply it to $\phi_2$ show no performance improvement. We believe that further improvements in CAD code editing will require larger and more diverse pre-trained CAD code generation datasets in the future.

\section{More Experimental Details}

\subsection{Baseline Implementation}
The prompt for close-source MLLMs (\ie, GPT-4o, Gemini 2.0, Claude 3.5) is presented in the table \ref{api_llm_prompt}.
Moreover, for a fair comparison, we also apply LoRA fine-tuning on the open-source Llama-3.2-11B model, with a LoRA rank of 8, a learning rate of 1e-5, 3 training epochs, a maximum sequence length of 8192, and a batch size of 2. 
For code editing, LoRA fine-tuning is conducted on another Llama 3.2 model with a LoRA rank of 64, 3 training epochs, a maximum sequence length of 8192 and a batch size of 1.

\subsection{Evaluation Metrics} \label{app_metric}
Following \cite{deepcad}, we adopt four common evaluation metrics to assess the quality of reconstructed 3D objects by edited CAD programs. 
While these metrics are originally designed for evaluating point cloud generation, we first convert these 3D objects into point clouds. 
These metrics are then computed by comparing a set of reference shapes $\mathcal{S}$ with a set of generated shapes $\mathcal{G}$. 
To better evaluate the outliers of reconstructed 3D objects, we report the mean chamfer distance (CD) scores in our experiments.

\noindent 
\textbf{Chamfer distance (CD)} is a metric used to evaluate the similarity between two point clouds, $P$ and $Q$.
Specifically, $P$ represents the reference point cloud (\ie, ground truth shape) and $Q$ represents the generated point cloud (\ie, reconstructed shape). 
It calculates the average squared distance from each point $p \in P$ to its nearest point $q \in Q$, and vice versa. 
The formula computes the minimum squared Euclidean distance $\|p-q\|^2$ for each point $p \in P$  to its closest neighbor in $Q$, and similarly for each point $q \in Q$ to its closest neighbor in $P$. 
CD can be defined as the sum of these minimum distances and normalized by the number of points in each set, providing a measure of how well the generated point cloud approximates the reference point cloud:
\begin{equation}
\begin{aligned}
\text{CD}(P, Q) = & \frac{1}{|P|} \sum_{p \in P} \min_{q \in Q} \|p - q\|^2 \\
& + \frac{1}{|Q|} \sum_{q \in Q} \min_{p \in P} \|q - p\|^2.
\end{aligned}
\end{equation}

\noindent \textbf{Minimum matching distance (MMD)} quantifies the fidelity of the generated shapes. It is calculated by determining the Chamfer Distance between each shape in the reference set $\mathcal{S}$ and its closest counterpart in the generated set $\mathcal{G}$. 
MMD can be defined as the average over all the nearest distances:
\begin{equation}
\text{MMD}(\mathcal{S},\mathcal{G}) = \frac{1}{|\mathcal{S}|}\sum_{Y \in \mathcal{S}}\min_{X \in \mathcal{G}}d^{CD}(X,Y).
\end{equation}

\noindent \textbf{Jensen-Shannon divergence (JSD)} is a statistical measure used to quantify the difference between two probability distributions. It is employed to assess the similarity between the reference set $S$ and the generated set $G$ by computing the marginal point distributions. 
The JSD can be mathematically expressed as: 
\begin{equation}
\begin{aligned}
\text{JSD}(P_S, P_G) = & \frac{1}{2} D_{\text{KL}}(P_S \parallel M) \\
& + \frac{1}{2} D_{\text{KL}}(P_G \parallel M),
\end{aligned}
\end{equation}
where $M = \frac{1}{2} (P_S + P_G)$. \( D_{\text{KL}} \) denotes the Kullback-Leibler (KL) divergence. The marginal distributions \( P_S \) and \( P_G \) are approximated by discretizing the space into 1024 voxel grids and assigning each point in the point clouds to one of these grids.

\noindent
\textbf{Invalid ratio (IR)} is calculated to measure the proportion of invalid CAD programs, which fail to compile for 3D object generation.

\begin{table*}[!]
    \centering
    \small
    \begin{spacing}{1.05}
    \caption{ \label{tab_feedback_prompt} The prompt for GPT-4o to annotate feedback on CAD programs.}
    \resizebox{\textwidth}{!}{
    \begin{tabular}{p{\linewidth}}
    \toprule[1pt]
    \textbf{Prompt}: 
        Please generate feedback for the 3D model based on the following information to support the code review process. \\
        - The first picture is the correct design drawing, and the second picture is the 3D model rendered by OpenSCAD code. \\
        - Error message: There is a Position anomaly on the code. \\
        - The length of comment: No more than 75 words. \\
        Requirements of Feedback Generation:\\
        1. First briefly describe the visual anomaly: clearly point out which component of the 3D model on the second picture is visually inconsistent with the design drawing. The description should focus on the deviation of the specific components in the model. \\
        2. Then describe the error in the code: explain the specific problem in the code that caused the visual anomaly. Do not give the correct code directly, but explain what is incorrect and point out the correct situation. \\
        3. Finally, point out the location of the erroneous code segment in the erroneous code to facilitate subsequent modifications. \\
        The erroneous code segment is as follows: \texttt{<erroneous\_code\_seg>} \\
        The correct code should be: \texttt{<correct\_code\_seg>} \\
        The full erroneous code is as follows: \texttt{<full\_erroneous\_code>} \\
    \midrule[1pt]
    \end{tabular}
    }
    \end{spacing}
\end{table*}

\begin{table*}[!]
    \centering
    \small
    \begin{spacing}{1.05}
    \caption{ \label{tab_cor_feedback} The list of predefined feedback for correct CAD programs.}
    \resizebox{\textwidth}{!}{
    \begin{tabular}{p{\linewidth}}
    \toprule[1pt]
    1. The 3D rendering captures the essence of the design blueprint with remarkable precision and fidelity. \\
    2. The 3D model matches the design drawing perfectly, with no deviations in key features like frames and recesses. \\
    3. The OpenSCAD-generated 3D model matches the original design drawing perfectly in all aspects.\\
    4. The 3D model mirrors the design drawing with exceptional clarity, maintaining all specified features.\\
    5. The implementation of the design in OpenSCAD results in a highly accurate and detailed 3D model.\\
    6. The alignment between the 3D rendering and the design drawing is precise, with all features correctly placed.\\
    7. The design intent is fully realized in the 3D model, with precise implementation of all structural elements.\\
    8. The faithful replication of the design drawing in the 3D model indicates precise coding and attention to detail.\\
    9. The correspondence between the design plan and the 3D model is seamless, with no misalignment or deviation.\\
    10. A careful analysis shows the 3D model to be a perfect reproduction of the design drawing.\\
    
    \bottomrule[1pt]
    \end{tabular}
    }
    \end{spacing}
\end{table*}

\begin{table*}[!]
    \centering
    \small
    \begin{spacing}{1.05}
    \caption{ \label{llm_aug_data_anno} The prompt for LLM-augmented data annotation for CAD localization.}
    \resizebox{\textwidth}{!}{
    \begin{tabular}{p{\linewidth}}
    \toprule[1pt]
    \textbf{Prompt}: Please generate a piece of OpenSCAD code that uses the boolean operations, such as \texttt{difference()}, \texttt{union()} and \texttt{intersection()} to construct a 3D model. The model should combine multiple geometric shapes, such as spheres, cubes, and cylinders to showcase the effect of the convex hull operation. Ensure that the internal component has a certain level of hierarchy and interlacing, and enhance the internal details. The code should be written in a modular manner for easy modification and adjustment, with clear readability and appropriate comments. \\
    \bottomrule[1pt]
    \end{tabular}
    }
    \end{spacing}
\end{table*}

\begin{table*}[!]
    \centering
    \small
    \begin{spacing}{1.05}
    \caption{ \label{api_llm_prompt} The prompt for closed-source MLLMs to perform the CAD review task.}
    \begin{tabular}{p{\linewidth}}
    \toprule[1pt]
    \textbf{Prompt}: 
        You are an expert in CAD design. 
You will be given correct reference images from multiple viewpoints and CAD program with the potential discrepancy, and your task is to identify the error type based on the given images and CAD program, write down the error, feedback and generate the correct code. \\
The error type must be one of the following: \\
No error: The code shown does not have any error \\
Missing block: A block is missing from the code \\
Redundant block: a block in the code is redundant \\
Size error: There is an error in the size of a block, e.g. the radius of a ball, the length of the sides of a cube. \\
Position error: A block has an incorrect translation or a wrongly added or missing translation. \\
Rotation error: The rotation angle of a block is incorrect or a rotation is incorrectly added or missing. \\
Primitive error: the shape of a block is incorrect, e.g. cube becomes cylinder or sphere, cylinder becomes cube or sphere, sphere becomes cube or cylinder
Logic error: an if or for condition error in the code. \\
Constant error: The value of a global variable in the code is wrong. \\
Requirements of feedback generation: \\
1. First briefly describe the visual anomaly: point out which component of the 3D model in the second image is visually inconsistent with the first image. The description should focus on the deviation of the specific components in the model. \\
2. Then describe the error in the code: explain the specific problem in the code that caused the visual anomaly and explain what is incorrect and point out the correct situation. \\
3. The feedback should be concise and clear, and no more than 75 words. \\
Example of feedback: The 3D model shows a deviation in the rotation of the component compared to the design drawing. Specifically, the rotation applied in Block 2 has an incorrect third parameter of 330 degrees instead of 0 degrees. This discrepancy results in the misalignment of the modeled part. Please correct the rotation in Block 2 to ensure proper orientation and alignment. \\
Requirement of code generation: \\
1. Based on the type of error and the feedback, generate a complete correct code in the same format as the erroneous code.\\
\textbf{Inputs}:
The input CAD program is: \texttt{<input\_code>}\\
The input reference images from multiple viewpoints are: \texttt{<reference\_images>}
Please output in JSON format, don't output anything else. The output format is as follows: \\
``error type'' : .., ``erroneous code block ID'': ..., ``feedback'' : ..., ``correct code'' : ... \\
    \bottomrule[1pt]
    \end{tabular}
    \end{spacing}
\end{table*}

\end{document}